\documentclass[a4paper]{article}
\usepackage{graphicx}
\usepackage{onecolpceurws}
\usepackage{booktabs}
\usepackage{amsmath}
\usepackage{amsfonts}
\usepackage{todonotes}
\title{Support Vector Machine Application for Multiphase Flow Pattern Prediction}

\author{
 Pablo Guill\'en-Rondon\\ University of Houston\\
                Houston, Texas 77004 \\ pgrondon@uh.edu
 \and
 Melvin D. Robinson \\University of Texas at Tyler\\
                Tyler, Texas 75579 \\ mrobinson@uttyler.edu
  \and
  Carlos Torres\\University of Los Andes\\Merida, 5101, Venezuela\\ctorres@ula.ve 
  \and
	\\Eduardo Pereya\\University of Tulsa\\Tulsa, Oklahoma 74104\\eduardo-pereyra@utulsa.edu
}
\institution{}

\begin{document}
\maketitle

\begin{abstract}
In this paper a data analytical approach featuring support vector machines (SVM) is employed to train a predictive model over an experimental dataset, which consists of the most relevant studies for two-phase flow pattern prediction. The database for this study consists of flow patterns or flow regimes in gas-liquid two-phase flow. The term flow pattern refers to the geometrical configuration of the gas and liquid phases in the pipe. When gas and liquid flow simultaneously in a pipe, the two phases can distribute themselves in a variety of flow configurations. Gas-liquid two-phase flow occurs ubiquitously in various major industrial fields: petroleum, chemical, nuclear, and geothermal industries.  The flow configurations differ from each other in the spatial distribution of the interface, resulting in different flow characteristics. Experimental results obtained by applying the presented methodology to different combinations of flow patterns demonstrate that the proposed approach is state-of-the-art alternatives by achieving 97\% correct classification. The results suggest machine learning could be used as an effective tool for automatic detection and classification of gas-liquid flow patterns.
\end{abstract}
\vskip 32pt
\section{Introduction}
Data mining, an essential step in the process of Knowledge Discovery in Databases (KDD), is defined as the automated extraction of patterns representing knowledge implicitly stored in large databases.  Data mining is not just a deductive database system or a statistical data analysis tool; it involves an integration of different techniques that rely on the interaction between users and data mining system.



One very practical application that has recently lent itself to the KDD space is detection of flow patterns.  Detection of flow patterns is a fundamental problem in two-phase flow analysis.  Two-phase gas-liquid flow occurs in many industrial applications and it is characterized by the simultaneous flow of gas and liquid through a pipe. When the gas and liquid flows in a pipe, the phases can arrange themselves in different configurations which are called flow patterns. These flow patterns play an important role on gas-liquid flow characterization. In general, pressure drop and liquid fraction are function of the flow pattern. These two parameters are essential for proper system design and troubleshooting. Nevertheless, other parameters are also flow pattern dependent. As an example, the wall shear stress and local velocities, which are required for corrosion and erosion calculations, highly depend on the flow pattern. 

Although gas-liquid flow has been widely studied, there is not a consensus on the number of flow patterns in two-phase flow due to overlapping and characterization subjectivity. This is especially true around the transition zones.   Shoham \cite{shoham2006mechanistic} attempted to summarize the main flow patterns for all inclination angles as dispersed bubble (DB), bubble (B), slug, churn (CH), annular (A), stratified smooth (SS) and wavy (SW). The flow pattern occurrence depends on operational conditions, pipe and fluid properties. Operational conditions are commonly represented by gas and liquid superficial velocities which corresponded to the in situ flowrate per unit of pipe area. Relevant pipe properties are the pipe inclination angle, inner diameter and roughness. Historically, surface tension, densities and viscosities of the phases are related with flow pattern occurrence.  Indeed all design variables, namely, phase velocity, pressure drop, liquid holdup, heat and mass transfer coefficients, residence time distribution, and rate of chemical reaction, are all strongly dependent on the existing flow pattern. Thus, knowledge of the existing flow pattern can help the industry carry out a better design of two-phase flow systems. These include accurate prediction of pressure drop and liquid inventory in pipe flow, and effective erosion corrosion planning, utilizing properly chemical additives, such as corrosion inhibitors and demulsifies \cite{pereyra2012methodology}. 
 
More generally multiphase flow, and specifically Multiphase Flow Metering (MFM), plays a key role in real-time monitoring of producing systems. Variables from MFM can be processed via data mining, and specifically the use of machine learning techniques can be used to estimate accurately flow patterns from data in fields with metering infrastructure \cite{al2018virtual}. Hence, the properties of multiphase flows within the reservoir and through the production network must also be taken into account for the production system to be fully characterized.  Recently, a new approach using artificial neural networks \cite{al2016artificial} and deep learning \cite{ezzatabadipour2017deep} has been applied for multiphase flow patterns prediction obtaining results with high precision of prediction.

The main objective of this work is to present a data analytics approach using an SVM algorithm, with inputs features to be used for different flow conditions such as pipe inclination and diameter, physical properties of the phases, and their superficial velocities, with the purpose of detect and classify a high percentage of flow patterns. The rest of this paper is organized as follows: Section 2 introduces the motivation for using support vector machines and explains our methodologies in creating the dataset.  followed by a description of the two-phase flow patterns database under study. Section 3 presents and discusses the results of the SVM model developed in this work and finally Section 4 presents some conclusions of this work.


\section{Materials and Methods}
\subsection{Support vector machines}
The SVM is a machine learning device that can be used for regression or classification \cite{vapnik2013nature}. The classifier separates data into two classes, but since all classification problems can be restricted to the consideration of the two-class classification problem without loss of generality, SVMs can be applied in multiclass setting. An SVM is trained by using training patterns to optimize its loss function by means of quadratic programming.  SVMs use kernel functions to achieve nonlinearity and high dimensionality to ultimately create linearly separable classes. 

Because the kernel function defines the feature space in which the training set examples will be classified, its selection is an important design consideration. Testing is based on the model evaluation using the support vectors to classify a test dataset. 
\subsection{Dataset}
A flow pattern experimental data base was collected \cite{pereyra2012methodology}, which consists of the most relevant studies developed in the area. Specifically for this study, the data set from Shoham \cite{shohamflow} was selected among the available sets due to its large number of data points (5676), range in inclination angle ($-90^\circ$ to $90^\circ$), two pipe diameters (ID=1in and 2in), and the wide range of flow patterns observed for all pipe inclination angles. The different flow patterns considered in this study are: 
\begin{itemize}
\item Annular flow (A): It is an segregated flow pattern where part of the liquid travels as film surrounding the pipe wall, the gas travels in the center of the pipe with some entrain liquid on it. This flow pattern occur at relative high gas superficial velocities.
\item Bubble (B): A multiphase fluid flow regime characterized by the gas phase being distributed as bubbles through the liquid phase.
\item Dispersed bubble (DB): This flow rate occurs at high liquid velocity where the turbulence is able to break the bubbles and distribute them uniformly along the pipe cross sectional area.
\item Intermittent (I): The intermittent pattern is usually subdivided into elongated bubble, slug and churn flows. Basically, these three flow patterns have the same configuration with respect to the distribution of the gas and the liquid interfaces. In these flow large bullet-shaped bubbles separate patterns slugs of liquid. In slug flow small gas bubbles aerate the liquid bridges. The elongated bubble pattern is considered the limiting-case of slug flow when the liquid slug is free of entrained bubbles \cite{barnea1985holdup}, while the churn flow takes place when the gas void fraction within the liquid slug reaches a maximum value above which occasional collapse of the liquid slug occurs. 
\item Stratified smooth (SS) and Stratified wavy (SW): Liquid and gas phases are completely separated and flow at low velocities.
\end{itemize}
The Intermittent flow pattern considers Slug (SL) and Churn (CH) flow pattern combined \cite{pereyra2012methodology}. In order to analyze the performance of the algorithm, three tests are proposed: Test 1 considers all the flow patterns proposed; Test 2 combines the SS and SW data points into stratified flow ST (ST = SS + SW); finally Test 3 combines the segregated flow patterns (ST + A) and the dispersed flow patterns (DB + B).
\subsection{Supervised classification of flow patterns}
In order to apply the support vector methodology for the classification of flow patterns a dataset for training and testing with classes (labels) was used. Three tests are proposed: Test 1 considers all the flow patterns proposed, this set considers 6 classes; Test 2 combines the SS and SW data points into stratified flow ST (ST = SS + SW), this set considers 5 classes; finally Test 3 combines the segregated flow patterns (ST + A) and the dispersed flow patterns (DB + B), this set considers 3 classes.
The predictor variables (input feature vectors) considered are the nine variables: velocity superficial liquid (VSL), velocity superficial gas (VSG), viscosity liquid, viscosity gas, density liquid, density gas, surface tension, inclination angle, and pipe diameters. We applied a conditioning (preprocessing) to the input features that standardize the data, giving each feature zero mean and unit variance.
\subsection{SVM Model parameter selection}
We will consider two parameters. 
\begin{enumerate}
\item The parameter $C$ is a regularization factor, and tells the classifier how much we want to avoid misclassifying training examples. A large value of $C$ will try to correctly classify more examples from the training set, but if $C$ is too large it may over-fit the data and fail to generalize when classifying new data. If $C$ is too small then the model will not be good at fitting outliers and will have a large error on the training set. 
\item The SVM learning algorithm uses a kernel function to compute the distance between feature vectors. In this study we are using a Gaussian radial base kernel function. The $\gamma$ parameter describes the size of the radial basis functions, which is how far away two vectors in the feature space need to be to be considered close.
\end{enumerate}

We will train a series of classifiers with different values for $C$ and $\gamma$. Two nested loops are used to train a classifier for every possible combination of values in the ranges specified. The classification accuracy is recorded for each combination of parameter values. The parameter values that give the best classification accuracy are selected.  In this way, the $\gamma$ value used was 10 for the three tests proposed, and the parameter $C=100$ was used for Test 1, and  $C = 1000$ for Test 2 and Test 3. We have applied our methodology and the results are presented in the next section.
\section{Results}
\begin{figure}
 \caption{Input features and flow patterns for the experimental dataset.}
 \label{fig:features}
 \includegraphics{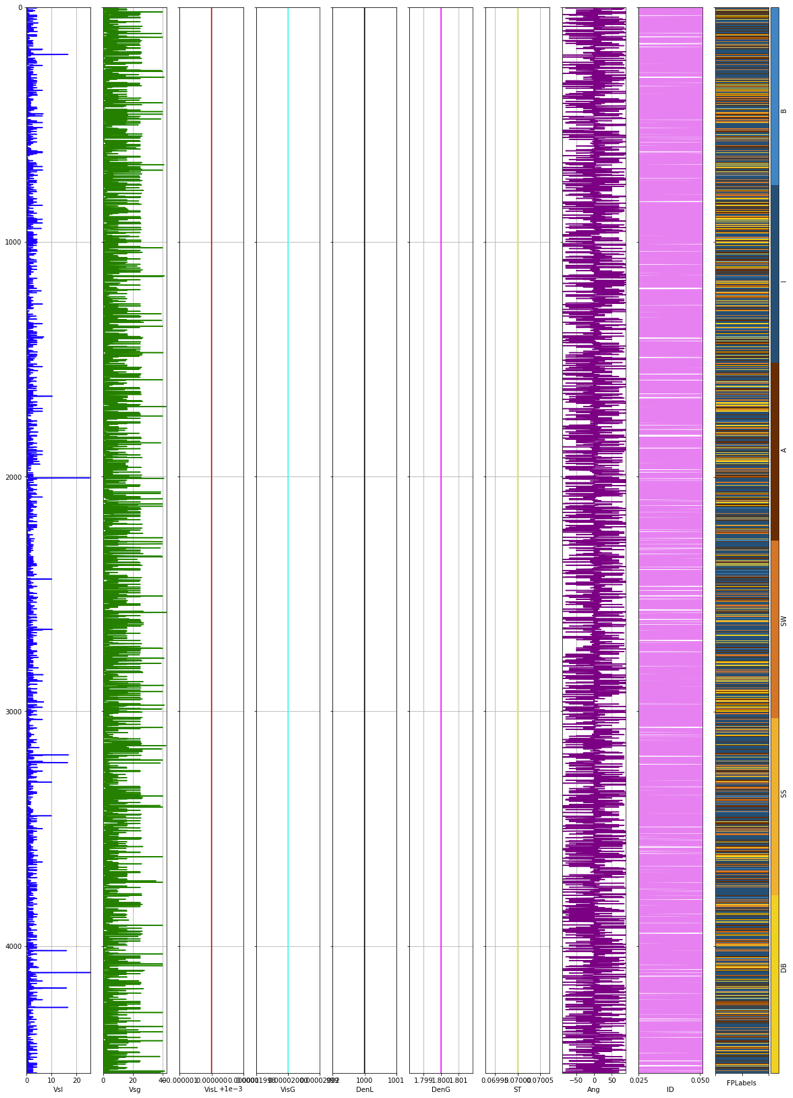}
 \end{figure}

In our approach, we trained a SVM algorithm on a set of randomly selected samples, approximately 80\% of the entire dataset was used for training, and approximately 20\% was used as the testing set. Fig. \ref{fig:features} shows the input features vectors (columns 1-9) and the flow patterns (FPLabels, last column) in the experimental dataset under study. 
 
In evaluating the effectiveness of our SVM methodology, the confusion matrix is an important measure. Table \ref{tb:test1} shows the confusion matrix for the training data for Test 1, predicting classes A, B, DB, I, SS, and SW. We can readily see the strong diagonal components. This means that our classifier is achieving little classification error.   
\begin{table}
\centering
\caption{Confusion matrix for the training data for Test 1.}
\label{tb:test1}
\begin{tabular}{ccccccc}\toprule
Flow Pattern&	DB &	SS   & 	SW     &	A &   I&     B\\\midrule
DB	&486&	0&	0&	0&	6&	0\\
SS	&0&	113&	0&	0&	0&	0\\
SW&	0&3&	675	&8&	0	&0\\
A&	0&	0&	12&	820&	1&	0\\
I&5&0&0&1&	2306&	0\\
B&	0&	0&	0&	0&	0&	104\\\bottomrule
\end{tabular}
\end{table}
\begin{table}
\centering
\caption{Classification results for the training data for Test 1.}
\label{tb:classification_t1}
\begin{tabular}{ccccc}\toprule
Flow Pattern	&Precision &     Recall &   $F_1$ &   Support\\  \midrule          
DB       &	0.99 &     	0.99 &      0.99  &        492           \\
SS       &	0.97    &  	1.00     &  0.99       &   113             \\
SW    &   	0.98     & 	0.98      & 0.98       &   686           \\
A     &  	0.99       &      0.98     &  0.99     &     833          \\ 
I     &  	1.00     & 	1.00    &   1.00      &   2312           \\
B     & 	1.00   &          1.00  &     1.00   &       104  \\
Avg/Total   & 0.99&             0.99 &      0.99  &        4540\\\bottomrule
\end{tabular}
\end{table}

Precision and recall are metrics that give more insights into how the classifier performs for individual flow patterns. Precision is the probability that given a classification result for a sample, the sample actually belongs to that class. Recall is the probability that a sample will be correctly classified for a given class. The $F_1$ score combines both to give a single measure of relevancy of the classifier results. Table \ref{tb:classification_t1} shows excellent results of these metrics for the training data. Practically, as shown in the confusion matrix and the relative statistics for each class, all classes had high rates of accuracy.  All these values would be considered excellent.

The testing set is used to predict the variable Flow Pattern, which contains labels for each class (A, B, DB, I, SS, and SW), and a predictive accuracy of 95\% for the different classes is obtained, the details of which are shown in Table \ref{tb:confusion_t1} and Table \ref{tb:class_test_1}. Off diagonal elements of a confusion matrix show misclassification with other flow patterns. The confusion matrix's columns represent the output patterns predicted by SVM while the rows represent the true class which is denoted here by each flow pattern.
\begin{table}
\centering
\caption{Confusion matrix for the testing data for Test 1.}
\label{tb:confusion_t1}
\begin{tabular}{ccccccc}\toprule
Flow Pattern&	DB&	SS	&SW	&A&	I&	B\\\midrule
DB&	95&	0&	0&	0&	7&	0\\
SS&	0&	27&	0&	0&	0&	0\\
SW&	0&	3&	179&	8&	2&	0\\
A&0&0&12&	180&8&	0\\
I&	8&	0&	1&	7&	577&	0\\
B&0&0&0&0&0&21\\\bottomrule
\end{tabular}
\end{table}

\begin{table}
\centering
\caption{Classification results for the testing data for Test 1.}
\label{tb:class_test_1}
\begin{tabular}{ccccc}\toprule
&                 Precision    &  Recall  &   $F_1$ & Support\\\midrule
DB    &   	0.92      &	0.93    &   0.93    &      102\\
SS       &	0.90      &	1.00    &   0.95      &    27\\
SW      & 	0.93      &	0.93     &  0.93       &   192\\
A       &	0.92           &  0.90     &  0.91     &     200\\
I      & 	0.97   &   	0.97   &    0.97     &     593\\
B     &  	1.00          &   1.00      & 1.00      &    21\\
Avg / Total &   0.95      &       0.95     &  0.95      &    1135\\\bottomrule
\end{tabular}
\end{table}

Similar to the discussion presented about the confusion matrix and the report of several metrics for the training data, we can see that for the testing data all class predicted had high rates of accuracy.

Confusion matrices obtained for training data considering 5 classes (Test 2) and 3 classes (Test 3) show accuracies of 99\%, similar to those shown in Table \ref{tb:test1}.

Table \ref{tb:confusion_t2} shows the confusion matrix on testing data for Test 2. Again as in Test 1, we can see the relatively strong diagonal components with noticeably lower off diagonal components. The testing set is used to predict the variable Flow Pattern, which contains labels for each class (A, B, DB, I, and ST), and we achieve a predictive accuracy of 95.00\%, the details of which are shown in Table \ref{tb:classification_t2}.  We can see that for the testing dataset all class predicted had high rates of accuracy.

\begin{table}
\centering
\caption{Confusion matrix for the testing data for Test 2.}
\label{tb:confusion_t2}
\begin{tabular}{ccccccc}\toprule
Flow Pattern	&DB&	ST	&A&	I&	B\\\midrule
DB&	185&	0&	0&	9&	6\\
ST&	0&	20&	0&	1&	0\\
A&	0&	0&	95&	7&	0\\
I&7&	0&	10&	573&	3\\
B&	11&	0&	1&	2&	205\\\bottomrule
\end{tabular}
\end{table}

\begin{table}
\centering
\caption{Classification Results for Test 2}
\label{tb:classification_t2}
\begin{tabular}{ccccc}\toprule
Flow Pattern	&Precision &     Recall &   $F_1$ &   Support\\\midrule             
A    &   	0.91     & 	0.93 &      0.92        &  102\\           
B      & 	1.00    & 	0.95   &    0.98        &  219  \\           
DB      & 	0.90     & 	0.93     &  0.91         & 200    \\       
I      & 	0.97     & 	0.97     &  0.97        &  593      \\     
ST     & 	0.96           &  0.94  &     0.95        &  21  \\
Avg /Total &   0.95  &           0.95 &      0.95 &         1135\\\bottomrule
\end{tabular}
\end{table}

Table \ref{tb:confusion_t3} shows the confusion matrix on testing data for Test 3 Again as in Test 1 and Test 2 we can see the relatively strong diagonal components with noticeably lower off diagonal components. The testing set is used to predict the variable Flow Pattern, which contains labels for each class (Intermittent, Dispersed, Segregate), and we achieve a predictive accuracy of 97.00\%, the details of which are shown in Table \ref{tb:results_t3}. We can see that for the testing data all class predicted had high rates of accuracy.


\begin{table}
\centering
\caption{Confusion matrix for the testing data for Test 3}
\label{tb:confusion_t3}
\begin{tabular}{cccc}\toprule
Flow Pattern	&Dispersed	&Segregate	&Intermittent\\\midrule
Dispersed	&575&	9&	9\\
Segregate	&7&95&	0\\
Intermittent&	12&	1&	427\\\bottomrule
\end{tabular}
\end{table}

 
\begin{table}
\caption{Classification Results for Test 3}
\label{tb:results_t3}
\centering
\begin{tabular}{ccccc}\toprule
Flow Patterns	&Precision &     Recall &    $F_1$ &   Support\\   \midrule
Intermittent   & 0.97         &  0.97 &    0.97      &  440 \\   
Dispersed      & 0.90     &      0.93  &   0.92     &   593  \\  
Segregate      & 0.98  &         0.97    & 0.97  &      102   \\
Avg /Total      & 0.97 &          0.97    & 0.97 &       1135\\\bottomrule
\end{tabular}
\end{table}

The results for the SVM approach for classification of two-phase flow pattern are encouraging. 

\section{Conclusions}
A data analytics approach using machine learning is proposed to quantify the prediction of gas-liquid two-phase patterns in pipes. Using an optimized SVM for the classification of flow patterns carries this out. We showed that the SVM algorithm could learn surprisingly well, as using our chosen kernel function and parameters allowed us to achieve high precision predicting different combinations of classes. When only three flow patterns are considered, namely, dispersed, segregated and intermittent, the predictive accuracy increases to 97\%. Finally, this study does provide some evidence that using machine learning techniques, as SVM is a promise mechanism to classify multiphase flow.

\section{Acknowledgements}
The authors would like to acknowledge CACDS for its support to present this work.

\bibliographystyle{alpha} 
\newcommand{\etalchar}[1]{$^{#1}$}

%
%
%

\end{document}